\documentclass[conference,compsoc]{IEEEtran}

\ifCLASSOPTIONcompsoc
  \usepackage[nocompress]{cite}
\else
  \usepackage{cite}
\fi
\usepackage{graphicx}
\usepackage{graphics}
\usepackage{amsmath}
\ifCLASSINFOpdf
\else
\fi

\begin{document}

\title{Indian Licence Plate Dataset in the wild}

\author{\IEEEauthorblockN{Sanchit Tanwar}
\IEEEauthorblockA{
Email: sanchittanwar75@gmail.com}
\and
\IEEEauthorblockN{Ayush Tiwari}
\IEEEauthorblockA{
Email: ayush123.at@gmail.com}
\and
\IEEEauthorblockN{Ritesh Chowdhry}
\IEEEauthorblockA{
Email: riteshchoudhery313@gmail.com}}

\maketitle

\begin{abstract}
Indian Number (Licence) Plate Detection is a problem that has not been explored much at an open-source level. There are proprietary solutions available for it, but there’s no big enough open-source dataset that can be used to perform experiments and test different approaches.
Most of the large datasets available are for countries like China\cite{xu2018towards}, Brazil\cite{gonccalves2018real}, but the model trained on these datasets do not perform well on Indian plates because the font styles and plate designs used vary significantly from country to country. 
This paper introduces an Indian Number (license) plate dataset with 16,192 images and 21,683 plate plates annotated with 4 points for each plate and each character in the corresponding plate. We propose a two-stage approach in which the first stage is for localizing the plate, and the second stage is to read the text in cropped plate image.  We present a model that uses semantic segmentation to solve number plate detection. We tested object detection and semantic segmentation model; for the second stage, we used lprnet\cite{zherzdev2018lprnet} based OCR. Data, codes and demo are available here: https://github.com/sanchit2843/Indian\_LPR

\end{abstract}

\IEEEpeerreviewmaketitle

\section{Introduction}
A critical feature to training a robust deep learning model is to ensure the dataset includes a wide variety of environments. Our dataset is developed with images captured in an unconstrained environment with varied camera types and viewpoints. As a result, a model trained with this dataset will be useful for real-world applications. 

\begin{figure}[h]
    \centering
    \includegraphics[scale=0.28]{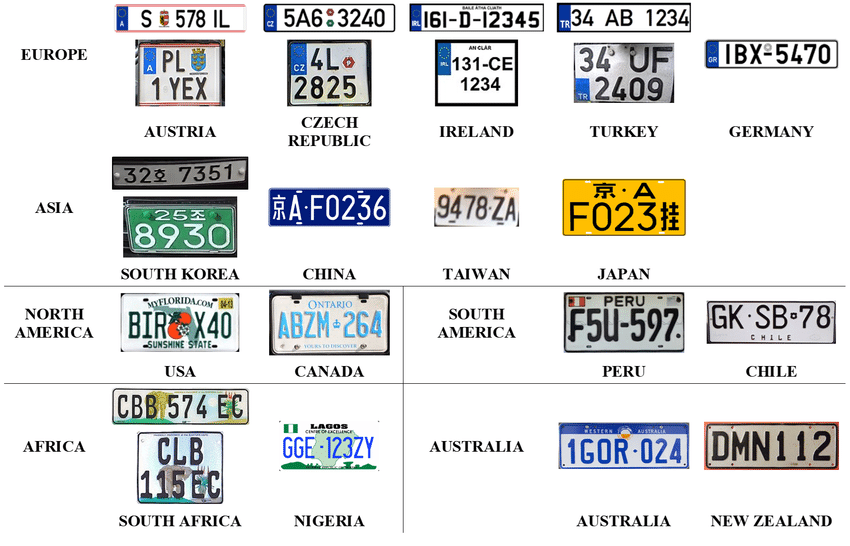}
    \caption{Different types of Licence Plates throughout the world \cite{differentPlates}}
    \label{fig:differentPlates}
\end{figure}

License plates vary from country to country in most cases. There are many unique number plate formats (as shown in fig \ref{fig:differentPlates}) throughout the world. Due to this, a model trained on one country's license plates does not perform well on recognizing text of another country's plates where this format is different.

Many open-source datasets already exist for Automatic Number Plate Recognition from China\cite{xu2018towards} and Brazil\cite{gonccalves2018real},
but to the best of our knowledge, there is no sizeable open-source dataset for Indian scenarios. To address the lack of data, we introduce an open-source dataset with 16192 images captured from road scenes and annotated with the location and characters of the number plates. We have described the dataset in detail in section \ref{sec:dataset} along with some examples, annotation details, and numbers.

To solve the number plate detection problem, the most common approach is to use object detection for the detection step, which essentially outputs a bounding box around a plate in an image. However, there are some problems with this approach which are discussed in section \ref{sec:model}. To mitigate those problems, we propose our benchmark model for this dataset, in which we have used semantic segmentation to do the detection step. The rest of the paper is divided into four sections: Related Work, Dataset, Model, Results, Conclusion, and Future Work.

\section{Related Work}

This section will briefly review some of the approaches used for solving automatic number plate recognition systems and other datasets available for number plate recognition. 
The problem of number plate recognition is usually solved in multiple stages, where the first stage is used to locate license plates and the second stage to recognize the characters of the plate from the cropped plate images found by stage 1. \cite{laroca2018robust} uses a similar approach in which yolov2\cite{redmon2017yolo9000} is used for detecting vehicles and number plates, for reading characters \cite{laroca2018robust} uses two separate networks where one is used for character segmentation and the second model is used for character recognition, where character segmentation model gives the bounding box proposals for the characters, these bounding boxes are used to crop the characters and fed into character recognition model. \cite{gonccalves2018real} uses a similar two-stage approach, for the second stage, the model uses seven fully connected layers as output, giving probabilities for the character at each location; such an approach can be used in settings with fixed-length output, but in Indian settings, the number of characters in plate varies. 
Some other methodologies use a single-stage approach and thus are end to end, \cite{xu2018towards} proposes a large dataset and a single-stage approach (RPNet) for solving number plate recognition. RPnet consists of one module for bounding box detection; these detections are then used to crop features from the feature maps using ROI Pooling\cite{girshick2015fast}, where these features are used with several classifiers to predict the license plate number. Thus this approach is unified and end-to-end trainable. Such an approach can also reduce the inference time as features are shared between detection and character recognition models.

To the best of our knowledge, all the approaches use object detection based methods for stage 1, the object detection poses challenges in cases of tilted number plates, where the cropped plates will have a lot of irrelevant information. \cite{zherzdev2018lprnet} uses spatial transformer network\cite{jaderberg2015spatial} to preprocess the cropped license plate image which helps in improving the results. One other solution can be to detect four separate points of plates, and use these 4 points with perspective transform to finally get a warped image. We have exploited this in our methodology by using semantic segmentation model for detecting polygon around plate and using perspective transform to get aligned plate as output, more details can be seen in section \ref{sec:model}.

Most datasets for License Plate Recognition usually collect images from traffic monitoring systems, highway toll station or parking lots. Caltech\cite{anagnostopoulos2008license} and Zemris\cite{hsu2012application} collected less than 700 images from high-resolution cameras on the road or freeways. CCPD\cite{xu2018towards} provides over 250k unique LP images with detailed annotations. All images are taken manually by workers of a roadside parking management company. The resolution of each image is 720 × 1,160. Each image in CCPD is labelled for license plate bounding box, four vertices location, horizontal and vertical tilt and other relevant information like the LP area. But each images in CCPD contains only a single license plate (LP). UFPR-ALPR\cite{laroca2018robust} dataset contains 150 videos and 4,500 frames captured when both camera and vehicles are moving and also contains different types of vehicles (i.e., cars, motorcycles, buses and trucks). Resolution of each image in UFPR-ALPR is 1,920 × 1,080. Every image has the following annotations available in a text file: the camera with which the image was taken, the vehicle’s position and information such as type (car or motorcycle), manufacturer, model and year; the identification and position of the LP, as well as the position of its characters.

\section{Dataset}
\label{sec:dataset}
The dataset contains a total of 16,192 images and 21,683 plates. The left table in the tables below (table \ref{tab:my_label}) shows the number of images of back and front plates, and the table on the right shows the number of types of vehicles in the images. There are images from different types of vehicles standard on Indian roads, which is essential because each vehicle type has a different plate placement, and in some cases, a different design and format can also be seen. 

\begin{table}[ht]
    \centering
    \begin{tabular}{|l|l|}
        \hline
        Back  & 17286 \\ \hline
        Front & 4397 \\ \hline 
    \end{tabular}
    \hspace{1cm}
    \begin{tabular}{|l|l|}
        \hline
        Car   & 10404 \\ \hline
        Bike  & 5393  \\ \hline
        Truck & 3284  \\ \hline
        3-Wheeler  & 1835  \\ \hline
        Bus   & 767 \\ \hline
    \end{tabular}
    \vspace{2pt}
    \caption{The table on the left shows the number of images of the back and front of vehicles, and the table on the right shows the number of images according to the type of vehicle.}
    \label{tab:my_label}
\end{table}

\begin{figure}[ht]
    \centering
    \includegraphics[scale=0.28]{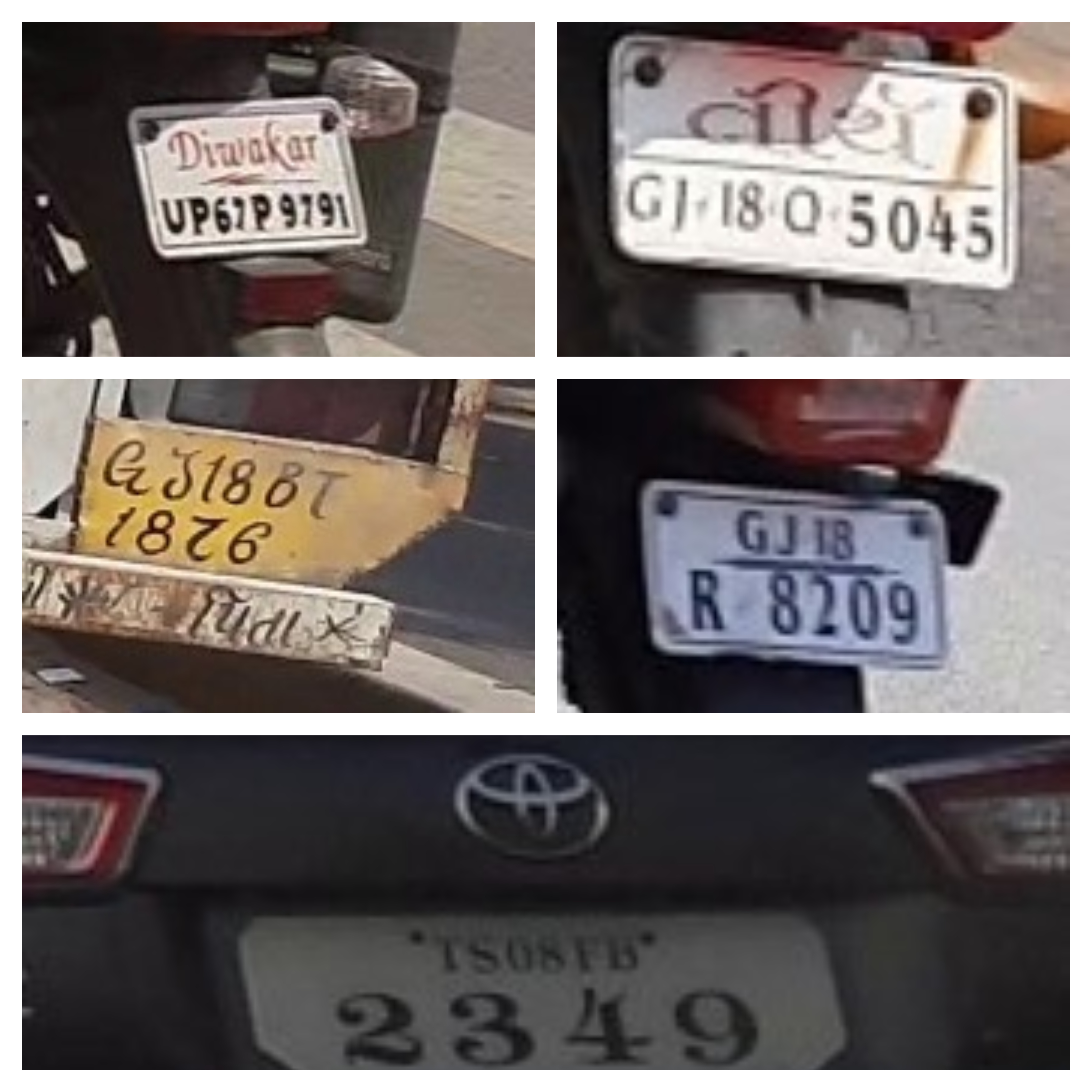}
    \caption{Some non-standard plate formats found in India}
    \label{fig:collage_varied_plates}
\end{figure}

\vspace{2pt}
\begin{table}[]
\begin{tabular}{|l|r|r|l|l|}
\hline
\textbf{Dataset} & \multicolumn{1}{l|}{\textbf{Year}} & \multicolumn{1}{l|}{\textbf{Images}} & \textbf{Resolution} & \textbf{Country} \\ \hline
Caltech Cars          & 1999 & 126                       & 896 ×592     & American      \\ \hline
EnglishLP             & 2003 & 509                       & 640 ×480     & European      \\ \hline
UCSD-Stills           & 2005 & 291                       & 640 ×480     & American      \\ \hline
ChineseLP             & 2012 & 411                       & Various      & Chinese       \\ \hline
AOLP                  & 2013 & 2,049                     & Various      & Taiwanese     \\ \hline
OpenALPR-EU           & 2016 & 108                       & Various      & European      \\ \hline
SSIG-SegPlate         & 2016 & 2,000                     & 1920 × 1080  & Brazilian     \\ \hline
UFPR-ALPR             & 2018 & 4,500                     & 1920  × 1080 & Brazilian     \\ \hline
CCPD                  & 2018 & \multicolumn{1}{l|}{250k} &              & Chinese  \\ \hline
Indian-LPR            & 2021 & \multicolumn{1}{l|}{16k}  & 1920 x 1080  & Indian        \\ \hline
\end{tabular}
\vspace{2pt}
\caption{This table shows the number of images in a few of the popular open source datasets in different countries along with the resolution of the images in the datasets (the last row is of our dataset).}
\label{tab:other_datasets}
\end{table}
There are images from a total of 10 states, they are filtered out from videos captured with different settings, devices, orientations and field of view. Some of the videos were captured from a dash cam placed inside a moving vehicle, some were captured with a person holding a camera on the road side, with the camera being static. Due to these differences in the video capturing, an unconstrained environment is created, meaning that the dataset has a huge variety in terms of type of camera being used, position of camera, viewpoint and thus it can be used in real-world applications. In the case of Indian Licence Plates there are many different non-standard formats that are widely used throughout India (refer fig \ref{fig:collage_varied_plates}) which creates issues with the character recognition, leading to inaccurate plate recognition. Although there's no way to cover all these images in a dataset, our dataset includes many images of this kind that are from vehicles on road in early 2021.

\subsection{Annotation}
\begin{figure}[ht]
    \centering
    \includegraphics[scale=0.09]{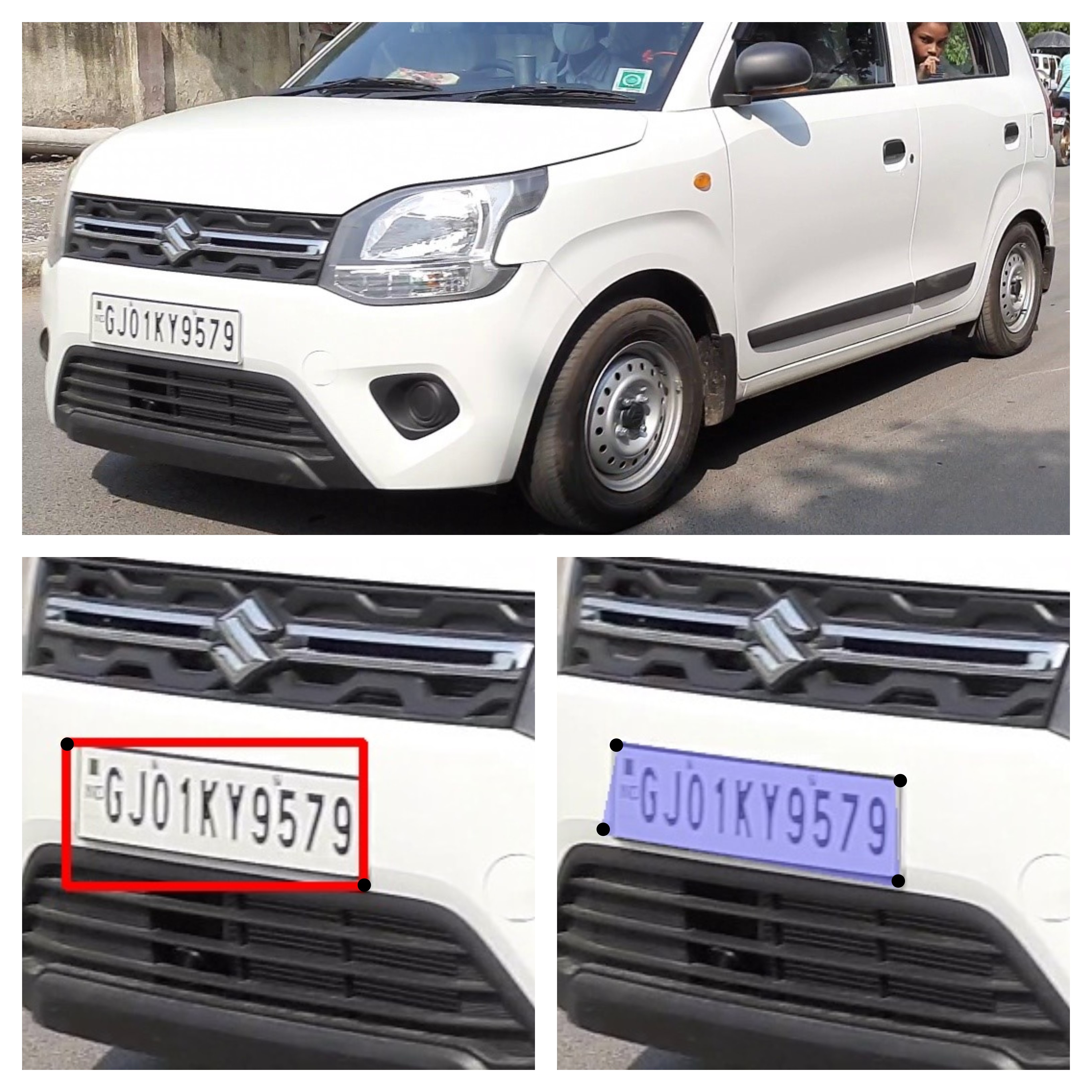}
    \caption{The benefit of using 4 point annotation, the upper image is the original image of the car, the bottom left is the 2 point annotation and the bottom right is the 4 point annotation}
    \label{fig:angled_plate_collage}
\end{figure}

The number plates observed by a camera on the road can be at a certain angle and orientation; in the case of a roadside camera, they are always at a certain angle, this leads to problems in identifying the plate in an image and performing OCR also becomes challenging. To tackle this problem and improve the robustness of OCR in the two-stage models, we provide annotation with 4 points around number plates (refer to figure \ref{fig:angled_plate_collage}) which provides orientation information as well, and thus we can use techniques like semantic segmentation, oriented bounding box detection for number plate detection. We can use the orientation information provided by the 4 points to straighten the number plate, eliminating any orientation issues for the model.

For each annotated plate we also provide information of the vehicle to which the plate belongs to and position of plate on vehicle (back or front), we also provide the characters of each plate for the character recognition part. The number plates which are completely readable are annotated, the unreadable or partially readable plates are not annotated.
\section{Model}
\label{sec:model}
We have created a baseline model with 2 stage approach, the first stage being the detection of number plates in an image, followed by a second stage to recognize characters in cropped plates. We tested two different models for the detection stage, first using object detection that has been well explored for the task of number plate detection in the past and then, a semantic segmentation-based model for number plate detection. Although, semantic segmentation is not a well-suited alternative for many object detection problems, because it does not work on the instance level but, for this problem it performs equally well since the individual instances (plates in this case) are well separated from one another and thus can be individually identified using simple post-processing steps. Although, the primary benefit of object detection over semantic segmentation is annotation time, but since for this task to annotate semantic segmentation polygons only 4 points are required (unlike the usual semantic masks which requires multiple vertices in the polygon), it is not much overhead over the basic rectangle-based object detection annotations. We crop the plates detected by any detection method and recognize them using LPRNet\cite{zherzdev2018lprnet} trained on our dataset. These models are individually described below.

Although we have provided the dataset with a division of the plates in 10 classes in total (car back, car front, bike around, bike front, and more), in this work, we trained the first stage model with a single class for the sake of simplicity.
\subsection{Object Detection}
We used a FCOS\cite{tian2019fcos} with Hrnet\cite{wang2020deep} backbone. FCOS is a single-stage object detection approach and thus works well in real-time applications such as ours. We used FCOS as the baseline model because FCOS is an anchor-free approach and requires minimal hyper-parameter tuning. The model was trained for 50 epochs with a batch size of 8 using ranger\cite{Ranger} optimizer. We used pixel-level augmentation techniques with random changes in brightness and contrast of the image. No spatial data augmentation was used in the baseline experimentation. The model was trained with a fixed resolution of 1920*1080 as downsampling the image can make the number plates unreadable. 

\subsection{Semantic segmentation}
We used similar Hrnet\cite{wang2020deep} backbone for semantic segmentation model. This backbone is followed by a convolution layer with output channels equal to the number of classes, i.e., 2 (the background and number plate). We used cross-entropy loss for training, with class weightage for background equal to 0.05 and 0.95 for number plate class because of the high level of class imbalance. We used Ranger\cite{Ranger} optimizer with 0.01 as the initial learning rate, and the learning rate was decayed using polynomial learning rate decay with degree 0.9. We trained the model for 50 epochs with a batch size of 8. Like object detection, only pixel-level augmentation techniques were applied, which randomly changed the image’s brightness and contrast. Image resolution for semantic segmentation was fixed at 1920*1080, and no cropping or downsampling was performed.

The output of semantic segmentation is polygonized. The polygon is then converted into a 4 point box; we use these 4 points to warp the plate as shown in Figure \ref{fig:angled_plate_collage}.

\subsection{LPRNET}
We used LPRNet\cite{zherzdev2018lprnet} for character recognition because it is a lightweight Convolutional Neural Network with high accuracy. We used LPRNet basic, which is based on Inception blocks \cite{szegedy2016rethinking} followed by spatial convolutions and trained with CTC loss. The model was trained for 100 epochs with Adam optimizer using a batch size of 32 with an initial learning rate of 0.001 and a gradient noise scale of 0.001. 
Data augmentation used were random affine transformations,e.g., rotation, scaling, and shift.

\subsection{Advantages of using Semantic segmentation}
Semantic segmentation has multiple advantages over object detection. The mask generated by semantic segmentation can be post-processed to generate 4 points for the plate, which has advantages over the 2 point method. 
It helped us crop out the plates irrespective of their orientation in the image, making the model usable for cases like roadside cameras.
\subsection{Failure mode of semantic segmentation}
\begin{figure}[ht]
    \centering
    \includegraphics[scale=0.3]{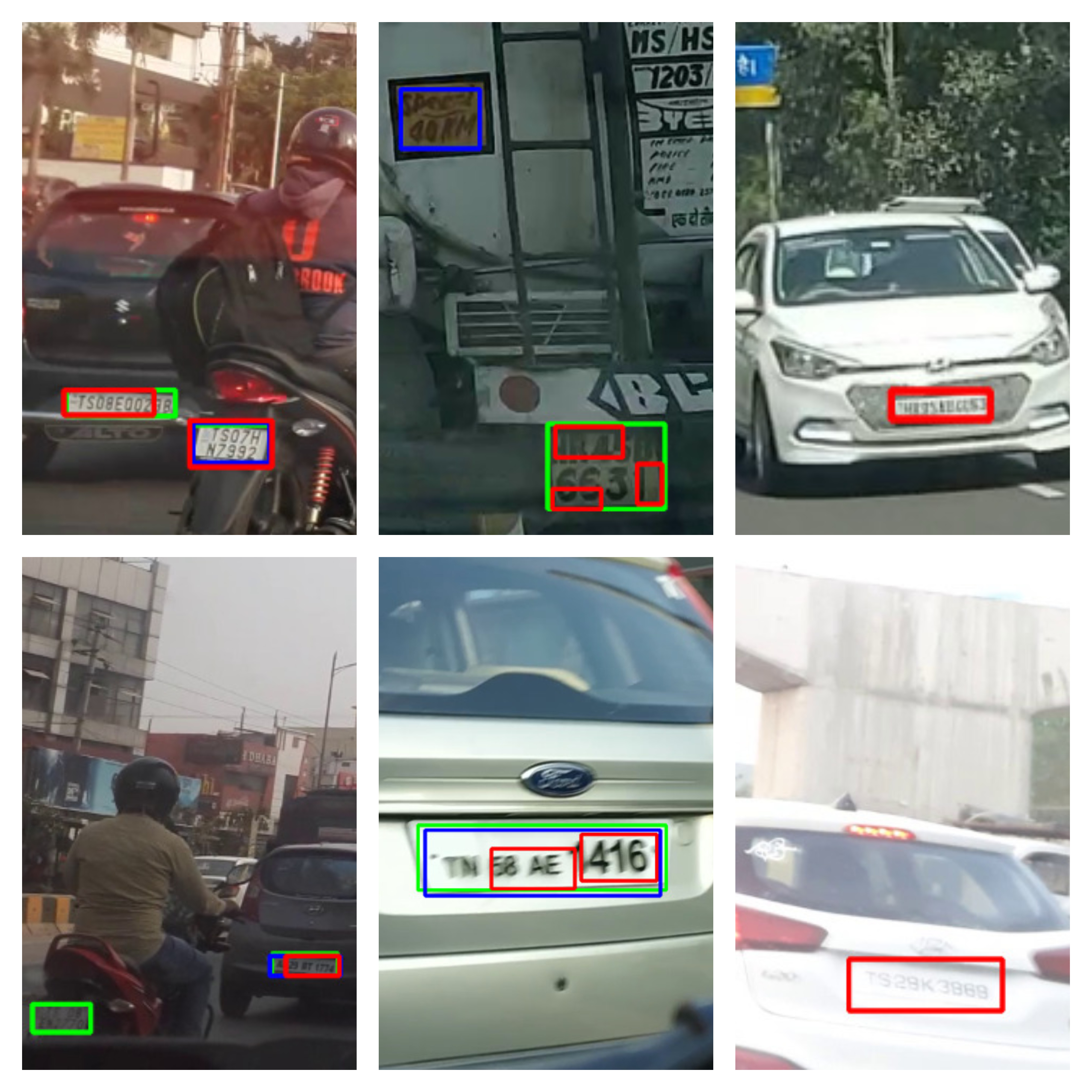}
    \caption{In the images, the green bounding box is ground truth; blue is object detection model output; red is semantic segmentation model output. The left column images show the incomplete bounding box failure mode of semantic segmentation. The middle column images show the bounding box split into multiple boxes failure modes. The third column images show the blurred plates being detected.}
    \label{fig:fail1}
\end{figure}

The semantic segmentation-based model also has a few limitations in this particular application; we found them frequently occurring during our testing. The examples of these failure modes are shown in figure \ref{fig:fail1}.

Here is a brief description of the failure modes which are shown in fig \ref{fig:fail1} above.
\begin{enumerate}
  \item The semantic segmentation model detected an incomplete bounding box small in size than the original plate and thus missing few characters, as shown in the two images in the first column. 
  \item The middle column images show examples in which a single plate is split into multiple smaller blobs, which usually happens with the plates in a 2-row format; this problem can be solved using convex hull-based post-processing or using dice loss instead of cross entropy might improve results in such scenarios.
  \item The third column images show the blurred plates which are detected by the semantic segmentation model but are not detected by the object detection model. 
\end{enumerate}

\section{Results}
In the tables below, we present the results of the model. We tested on the test split of the dataset, which has 3400 images; the fps was evaluated on an Nvidia 2070 super GPU at an image resolution of 1920*1080. The second table in table \ref{tab:results_tab} shows different cropping mechanisms that were used for the stage 2 model, where warp means taking perspective transform of the image using 4 points received from the semantic segmentation model, and thus the warped image will have pixels inside the 4 points of the original image. The rectangular crop means the crop with the rectangular bounding box received from the object detection model, this crop will have some extra information in the cropped image because of tilt that occurs naturally in the real world images.

\begin{table}[ht]
    
    \begin{tabular}{|l|l|l|}
        \hline
        Model & AP50   & FPS \\ \hline
        Object Detection  & 0.8310 & 12 \\ \hline
        Semantic segmentation  & 0.8313 & 14 \\ \hline
    \end{tabular}
    \centering
\vspace{5pt}

    \begin{tabular}{|l|l|}
        \hline
        Cropping Method  & Character Accuracy \\ \hline
        Warp & 75.6 \\ \hline 
        Rectangular Crop & 66.3 \\ \hline 
        
    \end{tabular}
    \vspace{2pt}
    \caption{Table on the left are the results of stage 1 model on test split, table on the right shows the character accuracy of LPRnet with different cropping methods.}
    \label{tab:results_tab}
\end{table}
\section{Conclusion}
To conclude, semantic segmentation model performs faster in comparison to object detection model and still achieves similar mAP. The advantage of using semantic segmentation model can be clearly observed in the results of stage two, where warping shows clear advantage of semantic segmentation based approach. Nevertheless, more work can be done on both types of model to achieve better results. The goal of this paper is to introduce a non-existing open source dataset for Indian licence plates, with the intention of promoting more experimentation.

\section{Future Work}
A semantic segmentation model can be tuned to improve the results; further, dice loss can be used to improve the results, especially in this problem because of the ability of dice loss to inherently handle class imbalance and being more susceptible to shapes. The character accuracy of the model in this work is 75\% which is not suitable for real-world scenarios; the primary reason for low accuracy is two-line plates which occur a lot in a real-world system. LPRnet fails to decode characters for two-row plates. A straightforward solution is to annotate the characters with rectangles around each character and using an object detection-based model in stage 2. But this requires a significant amount of annotation work, and thus there is a need for an OCR model that works well in two-row scenarios.

\bibliographystyle{IEEEtran}
\bibliography{bib.bib}

\end{document}